\documentclass[runningheads]{llncs}

\usepackage{eccv}

\usepackage{eccvabbrv}

\usepackage{graphicx}
\usepackage{booktabs}
\usepackage{multirow}

\usepackage[accsupp]{axessibility}  

\usepackage[pagebackref,breaklinks,colorlinks]{hyperref}

\usepackage{orcidlink}

\begin{document}

\title{StreetTree: A Large-Scale Global Benchmark for Fine-Grained Tree Species Classification} 

\titlerunning{StreetTree}

\author{Jiapeng Li\inst{1} \and
Yingjing Huang\inst{2} \and
Fan Zhang\inst{3} \and
Yu Liu\inst{4}}

\authorrunning{Jiapeng L., Yingjing H., et al.}

\institute{Peking University,
\email{jpli25@stu.pku.edu.cn}\\
\and
University of Vienna, 
\email{yingjing.huang@univie.ac.at}\\
\and
Peking University,
\email{fanzhanggis@pku.edu.cn}\\
\and
Peking University,
\email{liuyu@urban.pku.edu.cn}\\
}

\maketitle

\begin{abstract}
  The fine-grained classification of street trees is a crucial task for urban planning, streetscape management, and the assessment of urban ecosystem services. However, progress in this field has been hindered by the lack of large-scale, geographically diverse, and publicly available benchmark datasets specifically designed for street trees. To address this critical gap, we introduce StreetTree, the world’s first large-scale benchmark dataset dedicated to fine-grained street tree classification. The dataset contains over 12 million images covering more than 8,300 common street tree species, collected from urban streetscapes across 133 countries spanning five continents, and supplemented with expert-verified observational data. StreetTree poses challenges for pretrained vision models under complex urban environments: high inter-species visual similarity, long-tailed natural distributions, significant intra-class variations caused by seasonal changes, and diverse imaging conditions such as lighting, occlusions from buildings, and varying camera angles. In addition, we provide a hierarchical taxonomy (order–family–genus–species) to support research in hierarchical classification and representation learning. Through extensive experiments with various vision models, we establish solid baselines and reveal the limitations of existing methods in handling such real-world complexities. We believe that StreetTree will serve as a key resource for driving new advancements at the intersection of computer vision and urban science.
  \keywords{Urban forestry \and Fine-grained image classification \and Street view imagery \and Hierarchical taxonomy \and Long-tailed distribution}
\end{abstract}

\section{Introduction}
\label{sec:intro}

Urban street trees form an integral component of the global forest system, contributing substantially to carbon sequestration, biodiversity maintenance and local climate regulation~\cite{lian2025tree, wu2025tree}. 
Amid accelerating urbanization, these trees constitute the most visible and proximate form of urban greenery, directly influencing residents’ daily environments. 
They play critical roles in mitigating urban heat, improving air quality, and managing stormwater, thereby enhancing urban resilience and livability~\cite{smith2025integrated,schwaab2021role}. 
Establishing large-scale, high-quality, and scalable databases of street trees is thus fundamental for advancing evidence-based urban governance and supporting the planning and management of green infrastructure in cities worldwide.

However, constructing and maintaining such databases is both time- and labor-intensive, often relying on expert-led field surveys~\cite{beery2022autoa}. 
In recent years, remote sensing imagery and unmanned aerial vehicle (UAV) photography have been utilized to analyze tree species~\cite{bolyn2022mapping, thapa2024application}. 
Yet, these approaches are primarily suited to large suburban or forested areas, as their spatial resolution is typically too coarse to capture individual trees. 
Moreover, their top-down perspective provides only limited information on tree canopies, making fine-grained, species-level classification of individual trees particularly challenging.

The rapidly increasing amount of street view imagery has captured detailed visual information of urban streets~\cite{huang2023comprehensive, huang2025measuring}. 
It provides a comprehensive and in-depth description of urban environments, making it an excellent data source for observing street trees~\cite{wegner2016cataloging}. 
Drawing upon this rich data source, a growing body of research has used street view imagery to study urban street trees, including measuring urban greenery, identifying tree locations, and establishing tree inventories~\cite{huang2025no,arevalo-ramirez2024challenges,oliveira2022locating}.
While the importance of street view–based tree datasets has been increasingly recognized, several limitations remain. 
Existing datasets are still scarce, with limited geographic coverage and data scale, incomplete taxonomic annotations across hierarchical levels, noisy labeling quality, and often only single-time-point observations for each tree. 
These constraints continue to hinder the fine-scale management of urban street trees.

\begin{figure}[!h] 
  \centering
  \includegraphics[width=\linewidth]{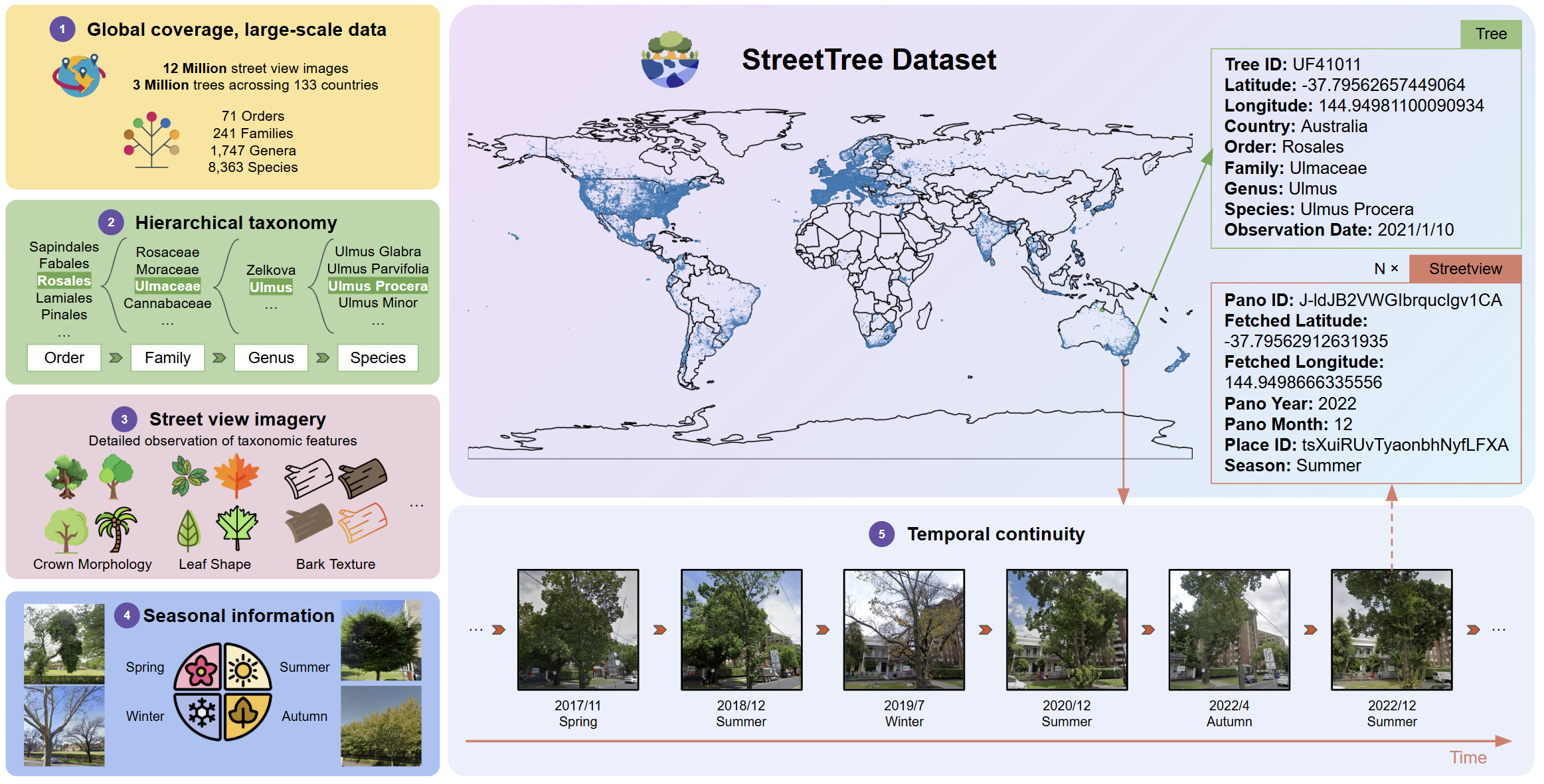}
  \caption{Overview of StreetTree dataset. The dataset comprises 12,235,152 street view images representing 3,365,485 individual urban trees across 133 countries and five continents, spanning 71 orders, 241 families, 1,747 genera, and 8,363 species. Each tree may correspond to one or multiple street view images. The map illustrates the global geographic distribution of tree samples, where darker shades indicate higher densities. Most individual trees are associated with long-term temporal records, capturing seasonal and interannual variations.}
  \label{intro}
\end{figure}

To address these limitations, we present \textbf{StreetTree}, a large-scale, fine-grained global dataset of urban street trees (Figure~\ref{intro}). The dataset includes samples from 133 countries across five continents, containing more than 12 million observations of 3,365,485 individual trees annotated with a four-level taxonomic hierarchy. Each tree in StreetTree is linked to its geographic coordinates, time and season of observation, and hierarchical taxonomic labels spanning order, family, genus, and species. Together, these components provide a comprehensive and structured representation of urban trees worldwide.

Overall, the StreetTree dataset offers five key contributions:
\begin{itemize}
    \item \textbf{Global coverage.} Over 12 million samples from 133 countries across five continents ensure broad spatial representativeness.
    \item \textbf{Hierarchical taxonomy.} Each tree is annotated with a four-level taxonomy: order, family, genus and species.
    \item \textbf{Street view imagery.} Ground-level images enable detailed observation of taxonomic features such as crown morphology, leaf shape, and bark texture.
    \item \textbf{Seasonal information.} Each tree is labeled with the corresponding season, supporting analysis of phenological variation.
    \item \textbf{Temporal continuity.} Spanning 2015–2025, the dataset provides long-term records for tracking urban vegetation change.
\end{itemize}

\section{Related Work}
\subsection{Publicly available tree species datasets}
Existing publicly available tree species datasets exhibit considerable diversity. 
While many are image-based, spanning from satellite to ground-level street view imagery, other tree inventories consist only of georeferenced and species data.
Table \ref{datasets} provides several representative datasets, detailing their geographic coverage, size, taxonomic granularity, temporal span, individual tree information, seasonal coverage, image type, and region type.

\begin{table*}[ht]
\centering
\caption{Overview of publicly available datasets for tree species classification. }
\label{datasets}
\resizebox{\textwidth}{!}{
\begin{tabular}{ccccccccc}
\toprule
\textbf{Dataset} & \textbf{\begin{tabular}{@{}c@{}}Geographic\\Scope\end{tabular}} & \textbf{Size} & \textbf{\begin{tabular}{@{}c@{}}Taxonomic\\Granularity\end{tabular}} & \textbf{\begin{tabular}{@{}c@{}}Temporal\\Span\end{tabular}} & \textbf{\begin{tabular}{@{}c@{}}Individual\\Tree\end{tabular}}  & \textbf{\begin{tabular}{@{}c@{}}Seasonal\\Coverage\end{tabular}} & \textbf{\begin{tabular}{@{}c@{}}Image\\Type\end{tabular}} & \textbf{\begin{tabular}{@{}c@{}}Region\\Type\end{tabular}} \\
\midrule
GlobalGeoTree~\cite{muGlobalGeoTreeMultiGranularVisionLanguage2025} & Global & 6,263,345 & Familiy / Genus / Species & 2015-2024 & Yes & No & Satellite Image & Forest \\ 
Auto Arborist~\cite{beery2022autoa} & USA & 2,637,208 & Genus & 2018-2022 & Yes & No & \begin{tabular}{@{}c@{}}Satellite Image and\\Street View Image\end{tabular} & Urban \\ 
U.S. 5 Million \cite{mccoySpeciesClusteringClimate2022} & USA & 5,660,237 & Genus / Species & - & Yes & No & Not available & Forest / Urban \\ 
Urbanforest~\cite{cityofmelbourne2023trees} & Australia & 76,928 & Family / Genus / Species & - & Yes & No & Not available & Urban \\
EU Forest~\cite{mauri2017euforest} & Europe & 588,983 & Genus / Species & - & No & No & Satellite Image & Forest \\ 
PureForest~\cite{gaydon2025pureforest} & France & 135,569 & Species & - & No & No & Satellite Image & Forest \\ 
TreeSatAI~\cite{treesatai} & Germany & 50,381 & Genus / Species & 2011-2020 & No & No & Satellite Image & Forest \\ 
\begin{tabular}{@{}c@{}}NEON Vegetation\\Structure Data~\cite{kampe2010neon}\end{tabular} & USA & - & Family / Genus / Species & - & No & No & Satellite Image & Forest \\  
\midrule
\textbf{StreetTree} & \textbf{Global} & \textbf{12,235,152} & \textbf{\begin{tabular}{@{}c@{}}Order / Family /\\Genus / Species\end{tabular}}& \textbf{2015-2025} & \textbf{Yes} & \textbf{Yes} & \textbf{Street View Image} & \textbf{Urban} \\
\bottomrule
\end{tabular}
}
\end{table*}

Despite their considerable value, these existing datasets still present several shortcomings that remain to be addressed. First, most datasets are limited to a single country or city, resulting in extremely narrow geographic coverage that restricts data diversity and hampers the transferability of models across regions. Second, few datasets provide a complete taxonomic hierarchy, making it difficult to satisfy taxonomic query requirements in a unified framework. Third, some datasets only offer coarse geographic information (e.g. EUForest~\cite{mauri2017euforest}, Pureforest~\cite{gaydon2025pureforest}, TreeSatAI~\cite{treesatai} and NEON Vegetation Structure Data~\cite{kampe2010neon}), such as approximate locations within $1 km \times 1 km$ grids, rather than coordinates of individual trees, which prevents accurate spatial localization of species. Last, all these datasets lack long-term observations and seasonal information at the individual-tree level. Although certain datasets span multiple years, each tree is typically recorded only once, limiting the ability to continuously monitor tree growth and phenological dynamics.
Therefore, we introduce the StreetTree dataset to fill these gaps.

\subsection{Applications of street view imagery in urban forestry}
Owing to its ability to provide fine-grained, eye-level information on trees at the street scale, street view imagery has increasingly emerged as a crucial data source for urban forestry research.
Specifically, researchers use this direct street-level view by applying computer vision techniques such as semantic segmentation and object detection.
These methods facilitate various applications, including quantifying urban greenery~\cite{huang2025no}, assessing biodiversity~\cite{velasquez-camacho2024assessing}, and detecting pests and diseases~\cite{kagan2021automatic}. These results have also been utilized to further examine the role of vegetation in mitigating the urban heat island effect and improving street-level thermal environments~\cite{huang2025exploring, li2022exploring}.

Fine-grained tree species classification using street view imagery is a particularly complex task. 
Although pioneering work has demonstrated the feasibility of this approach~\cite{branson2018google, wang2025urban}, the majority of existing studies are confined to a limited set of species or small geographic regions, hindering the generalizability of their findings. 
While some efforts have aimed to develop larger-scale datasets to propel this field forward~\cite{beery2022autoa, mccoySpeciesClusteringClimate2022}, these resources still lack the necessary scale and comprehensiveness to adequately support the fine-grained classification and management demands of urban forestry.

\section{StreetTree Dataset}
\subsection{Dataset construction}
The StreetTree dataset provides a set of street view imagery and multi-level taxonomic annotations for individual urban street trees worldwide. The dataset construction process consists of two stages. First, we collect and preprocess the taxonomic and geolocation data. Then, these geolocations are used to query, download and preprocess the corresponding street view imagery.

\paragraph{\textbf{Taxonomic and geolocation data}}
To achieve broad geographic and taxonomic coverage, we aligned and integrated data from multiple sources. 
These sources included recently published academic datasets (e.g., TreeML~\cite{treemldata}, U.S. 5Million~\cite{mccoySpeciesClusteringClimate2022}, and UrbanForest~\cite{cityofmelbourne2023trees}), online resources (e.g., Singapore Trees~\cite{ExploreTreesSG}), data applied from governmental agencies (e.g., Cambridge Tree Dataset), and the Global Biodiversity Information Facility (GBIF~\cite{GBIF_occurrence}).

The GBIF dataset, in particular, required a specialized collection and filtering process due to its comprehensive global scale. 
To effectively query this extensive dataset, we first constructed a global tree species catalog based on two major inventories, TreeGOER~\cite{kindt2023treegoer} and GlobalTreeSearch v1.9~\cite{beech2017globaltreesearch}, which includes 59,417 unique species. 
Subsequently, we used this catalog to query the GBIF database. 
To ensure the quality and reliability of the retrieved GBIF records, we applied a series of filtering criteria. 
We restricted the temporal range to observations recorded between 2015 and 2025 and limited the data source to human observations, minimizing interference from automated detections or specimen-based records. 
Furthermore, we excluded records with geospatial issues or inconsistencies, such as missing coordinates or mismatched country information, and retained only those records with an occurrence status of “present”, excluding any trees marked as dead or absent.

After compiling the filtered data from all sources, we refined the consolidated dataset to focus exclusively on urban street trees. 
This was achieved through an overlay analysis using the Global Human Settlement Layer (GHSL)~\cite{ghsl} data, which delineates built-up areas globally.
We identified and retained only those trees located within these urbanized zones, ensuring the dataset focuses on urban environments. 
Finally, for records with incomplete taxonomic information, we further utilized our tree species catalog to impute the missing multi-level taxonomic labels. 
To resolve taxonomic synonyms and ensure Darwin Core compliance, we aligned all raw labels with standard GBIF species entries via the GBIF Taxonomic Backbone. Our labels were also verified against government data and institutional archives (e.g., museums), and we manually validated a random subset to guarantee high fidelity. 
For trees with a missing observation date, the publication date of the corresponding dataset was used instead.

\paragraph{\textbf{Street view imagery}}
Using the geographic coordinates from the tree records, we queried the Google Street View (GSV) API to identify the nearest sampling point for each tree. 
We then retrieved all available temporal images associated with that sampling point, as a single location often provides imagery from multiple collection dates. 
Crucially, to ensure each image was correctly centered on the target, we computed the azimuth angle from the capture point toward the tree location and used this orientation to download the street view image.

\begin{figure*}[htbp]
  \centering
  \includegraphics[width=\textwidth]{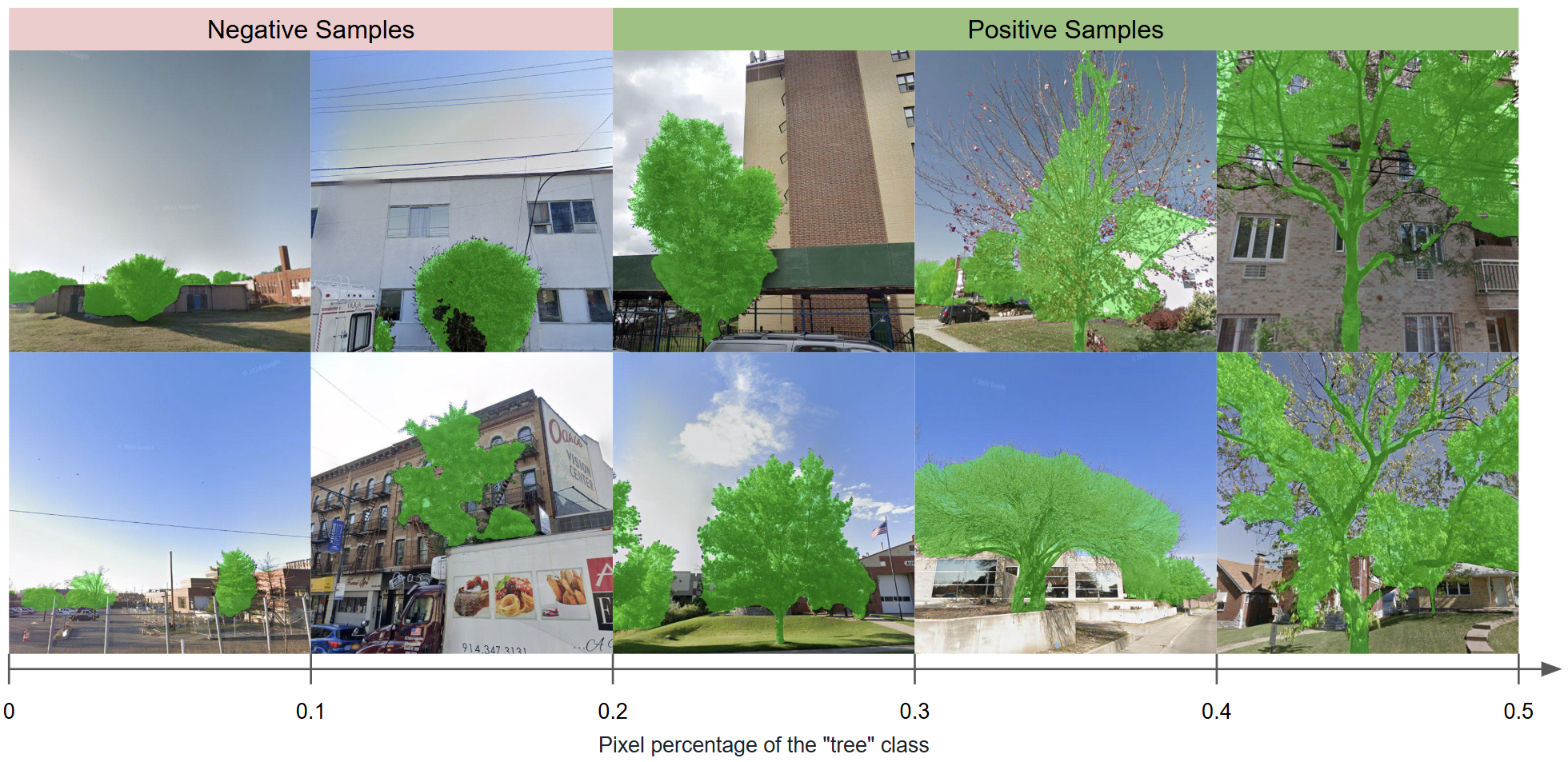}
  \caption{Examples of the binary classification of street view images. Images with less than 20\% “tree” class pixels are labeled as negative samples, whereas those exceeding 20\% are labeled as positive samples. Segmentation masks of “tree” class are highlighted with green color.}
  \label{binary}
\end{figure*}

Uncertainties during the GSV collection process can lead to invalid images, which may suffer from issues such as poor illumination (low or excessive), occlusion by buildings or vehicles, or camera angle deviations.
These issues often result in the target tree being missing or blurred in the image.
To filter out such invalid samples, we designed a two-stage filtering procedure.
In the first stage, we randomly sampled 10,000 images from the entire street view dataset and applied the Segformer~\cite{xie2021segformer} for semantic segmentation to compute the pixel percentage of the “tree” class. 
An image was classified as a negative sample (tree absent or unclear) if this percentage was below 20\%, and as a positive sample otherwise (see Figure~\ref{binary}).

In the second stage, we trained an efficient binary classifier to automate this filtering process for the entire dataset.
To enhance classification efficiency, we did not use the raw images; instead, we first extracted deep features using a ResNet18~\cite{he2016deep} model pretrained on the Places365~\cite{zhou2018places} dataset.
These features were then used to train the classifier, which consisted of a two-layer fully connected multilayer perceptron (MLP).
The model was trained for 30 epochs with a batch size of 4096 and a learning rate of 1e-3, using a 9:1 training-validation split. 
On the validation set, it achieved accuracy of 0.9590 and recall of 0.8214. 
Finally, this binary classifier was applied to predict and filter the remaining street view images, retaining only the positive samples.
Manual inspection confirmed that sub-threshold images suffer from severe occlusion or lack discriminative texture details for classification; excluding these noisy samples prevents feature learning degradation and ensures high training efficacy.

\subsection{Dataset description}
The StreetTree dataset contains 12,235,152 samples representing 3,365,485 individual urban street trees across 133 countries. 
Each record consists of three key components: (1) the taxonomic and spatial attributes of the tree; (2) the corresponding street view image; and (3) the metadata associated with that image.
Specifically, the taxonomic and spatial attributes of the tree include its geographic coordinates (latitude and longitude), its location (country and city), the observation date, and a four-level taxonomic hierarchy (order, family, genus, and species; e.g., Order: Sapindales; Family: Sapindaceae; Genus: Acer; Species: Acer rubrum).
The street view image metadata includes the latitude and longitude of the GSV sample point, the year and month of image capture, and the corresponding season.

\begin{figure}[htbp]
  \centering
  \includegraphics[width=0.8\linewidth]{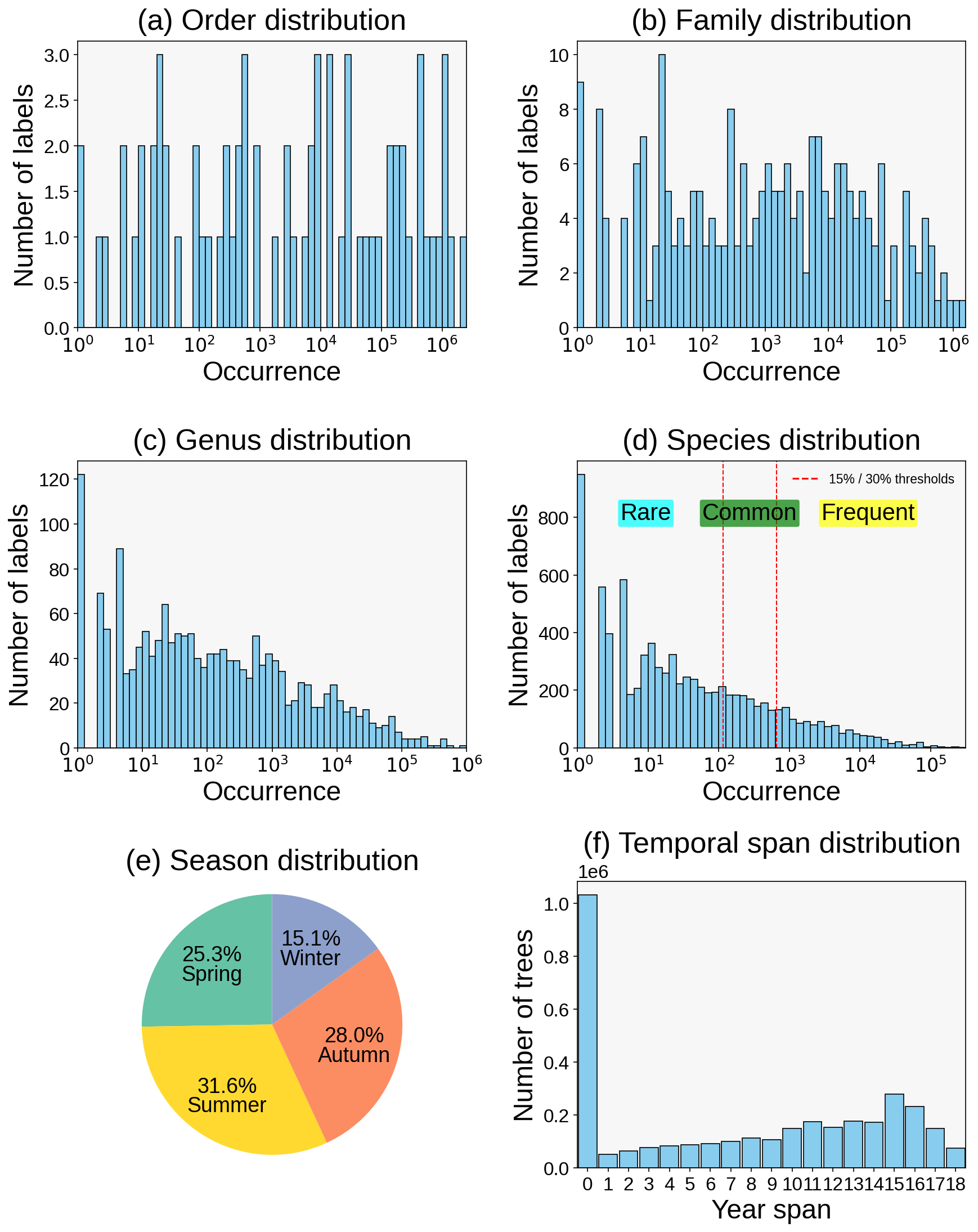}
  \caption{Statistics of StreetTree dataset. \textbf{(a)–(d)} show distributions across the four taxonomic levels: Order, Family, Genus, and Species. The horizontal axis uses a logarithmic
 scale to represent the magnitude of occurrence counts; \textbf{(e)} presents the seasonal distribution; \textbf{(f)} illustrates the temporal span distribution of individual trees.}
  \label{stats}
\end{figure}

We analyze the key distributional characteristics of the StreetTree dataset, focusing on its taxonomic composition, seasonal distribution, and temporal coverage.
As can be seen from Figure~\ref{stats} (a)-(d), the sample distribution exhibits a pronounced long-tailed structure across all four taxonomic levels (order, family, genus, and species).
Statistical indicators show that at the genus level, skewness is 12.27 and kurtosis is 198.46; at the species level, skewness reaches 14.71 with a kurtosis of 292.03. 
These values reflect extremely strong positively-skewed and heavy-tailed distributions, in which a small number of high-frequency classes account for a large proportion of samples, while the vast majority of classes appear only rarely. 
This structure mirrors the inherent unevenness of real-world urban tree distributions and implies that models may naturally bias toward frequent classes while facing generalization challenges on rare categories.

For the seasonal distribution, Figure~\ref{stats} (e) show a relatively balanced yet slightly uneven seasonal composition. 
Spring, summer, autumn, and winter account for 25.3\%, 31.6\%, 28.0\%, and 16.1\%, respectively.
This distribution reflects the phenological cycles throughout the year and ensures that the dataset captures substantial seasonal variation in tree appearance, including leaf color, bark texture, and growth status. 
Overall, the diversity of seasonal observations enables models to learn season-dependent visual patterns more effectively and enhances their ability to adapt to appearance variation in real-world environments.

Regarding the temporal distribution, StreetTree demonstrates substantial long-term observational depth in Figure~\ref{stats} (f). 
A total of 921,723 trees have observation records spanning from 1 to 10 years, and 1,411,615 trees have records spanning from 11 to 18 years. 
Even trees without multi-year histories may include observations from different seasons within a year. 
This temporal span not only covers a wide range of longitudinal time scales but also provides valuable historical records for studying long-term patterns of urban tree growth dynamics.

Together, these three complementary dimensions reveal the richness and complexity of the StreetTree dataset’s statistical structure: a naturally long-tailed distribution in the species dimension, pronounced intra-species morphological variation across seasons, and extensive long-term observational depth over time. 
These characteristics capture the complexity of urban ecological systems and offer diverse and meaningful signals for visual learning across multiple scales, while simultaneously imposing higher demands on model design, class-balancing strategies, and generalization ability.

\subsection{Dataset partitioning}
To enable effective model training and evaluation, the StreetTree dataset was divided into two parts: the training set \textbf{StreetTree-12M} and the evaluation set \textbf{StreetTree-18kEval}.

\textbf{StreetTree-12M} contains the vast majority of samples and is used for fine-tuning large-scale vision models such as Vision Transformer (ViT)~\cite{dosovitskiy2020image}, CLIP~\cite{radford2021learning}, SigLIP~\cite{zhai2023sigmoid} and BioCLIP~\cite{stevens2024bioclipa}. The extensive volume of data allows these models to learn rich visual and contextual representations of tree species and their surrounding urban environments.

\textbf{StreetTree-18kEval} serves as a benchmark designed to assess model performance in a fair and robust manner across both taxonomic levels and species-frequency categories. To address the long-tailed distribution of species, all species were ranked by their frequency of occurrence and grouped into three categories: frequent, common, and rare. Specifically, frequent species correspond to the top 15\% of species by occurrence frequency, common species represent the next 15–30\%, and rare species include the remaining 70\% (see Figure \ref{stats} (d)).
In this evaluation set, 30 species were randomly selected from each category. For each frequent species, 500 samples were drawn; for each common species, 100 samples; and for each rare species, 20 samples, resulting in a total of 18,600 images.
To prevent potential data leakage where different images of the same individual tree might accidentally appear in both the training and evaluation sets, leading the model to memorize specific instances rather than learn generalizable species features, we enforced a strict instance-level separation. For the evaluation of models trained on varying data proportions, we dynamically filtered the base \textbf{StreetTree-18kEval} set. Any test images belonging to individual trees present in the training subset were systematically excluded. To preserve the size and balanced distribution, these excluded samples were replaced with images of different individual trees from the corresponding species. Its geographic distribution closely aligns with that of the full training dataset, broadly spanning multiple regions around the globe and ensuring strong global representativeness.

\section{Benchmarks}
\subsection{Baseline models and experimental setup}
To evaluate the performance of current mainstream model architectures in large-scale tree species classification, we compared three models that differ in both training set and architectural design: ViT fine-tuned on StreetTree-12M, CLIP pretrained by OpenAI , CLIP fine-tuned on StreetTree-12M, SigLIP fine-tuned on StreetTree-12M and BioCLIP fine-tuned on StreetTree-12M. 
We aim to establish a benchmark for global urban street tree species classification in order to reavel the limitations and potential improvements of existing architectures, thereby enhancing our understanding of urban tree biodiversity across the world.

\paragraph{\textbf{Fine-tuned ViT}}
We initialized the ViT model with weights pretrained on ImageNet~\cite{deng2009imagenet} and then fine-tuned it on the StreetTree-12M dataset.
Specifically, all parameters of the backbone were unfrozen during this process, and the model was trained to simultaneously predict the four hierarchical taxonomic levels: order, family, genus, and species.


\paragraph{\textbf{CLIP}}
We used the CLIP visual encoder as a feature extractor for the StreetTree-12M dataset. 
The entire CLIP backbone was kept frozen during training. 
Only the classification heads were trained to predict the four hierarchical taxonomic levels: order, family, genus, and species. 

\paragraph{\textbf{Fine-tuned CLIP / Fine-tuned SigLIP / Fine-tuned BioCLIP}}
To explore the efficacy of cross-modal alignment, we fine-tuned several CLIP-based architectures.
First, we fine-tuned both the visual and text encoders on the StreetTree-12M dataset, using the images and their corresponding hierarchical textual labels. 
Then, these fine-tuned encoders were frozen. 
The visual encoder was then used as a feature extractor to train four classification heads for the hierarchical taxonomic levels (order, family, genus, and species).


\paragraph{\textbf{Data efficiency experiment}}
To evaluate the data utilization efficiency of different models, we designed a data efficiency experiment. 
While the total size of the StreetTree-12M set reaches tens of millions of samples, we hypothesize that competitive performance can still be achieved even with a small fraction of the data. 
Specifically, we constructed four training subsets containing 0.1\% (12K samples), 1\% (120K samples), 10\% (1.2M samples), and 100\% of the StreetTree-12M dataset, respectively. 
To account for sampling variability, each of the three smaller subsets (0.1\%, 1\%, and 10\%) was independently sampled five times using different random seeds.
This experiment analyzes how model performance varies with the amount of available training set under identical conditions, thereby allowing us to assess the robustness and generalization capability of different models under data-scarce scenarios.

\paragraph{\textbf{Training details}}
All models were trained using PyTorch~\cite{paszke2019pytorch}. 
The ViT model was fine-tuned for 10 epochs on four NVIDIA L20 GPUs (48 GB). 
We used the Adam optimizer~\cite{loshchilov2019decoupled} with a learning rate of $5 \times 10^{-4}$ and a total batch size of 256. 
For the CLIP, SigLIP and BioCLIP fine-tuning, we used four NVIDIA RTX 4090 GPUs (24 GB) and the AdamW optimizer (weight decay 0.01) with a per-GPU batch size of 128. 
The first stage (contrastive learning) ran for 5 epochs with a learning rate warm-up from $1 \times 10^{-5}$ to $2.5 \times 10^{-4}$. 
The second stage (classifier training) ran for 10 epochs at a constant learning rate of $2.5 \times 10^{-4}$. 
Both full training runs required approximately two days to complete.

\subsection{Experimental results and evaluation}
For all models, predictions were performed on the StreetTree-18kEval benchmark, dynamically adjusted for each training subset to maintain strict instance-level separation. Model performance was assessed using Top-1 and Top-5 accuracy across the four taxonomic levels (order, family, genus, and species). To offer a granular understanding of how class imbalance affects learning, the mean accuracies are systematically grouped by species-frequency categories (\textit{frequent}, \textit{common}, and \textit{rare}) and detailed in Table~\ref{result}. Additionally, the overall mean accuracies and standard deviations across 5 random runs are summarized in Table~\ref{result_all} to quantify general model performance and stability. Note that only a single run was performed for models trained on the full training set.

\begin{table*}[ht]
\centering
\caption{Model evaluation on StreetTree-12M. Results are presented as mean accuracy (\%) over 5 random runs.}
\label{result}
\small
\setlength{\tabcolsep}{4pt}
\renewcommand{\arraystretch}{1.1}
\resizebox{\textwidth}{!}{
\begin{tabular}{c*{12}{c}}
\toprule
\multirow{2}{*}{\textbf{Model}} & \multicolumn{4}{c}{\textbf{Frequent}} & \multicolumn{4}{c}{\textbf{Common}} & \multicolumn{4}{c}{\textbf{Rare}} \\
\cmidrule(lr){2-5}\cmidrule(lr){6-9}\cmidrule(lr){10-13}
 & Order & Family & Genus & Species & Order & Family & Genus & Species & Order & Family & Genus & Species \\
\midrule
\multicolumn{13}{l}{\textbf{0.1\% of training set}} \\
\midrule
Fine-tuned ViT & 17.52 (52.31) & 13.26 (34.56) & 6.33 (20.16) & 1.81 (5.76) & 12.34 (38.11) & 9.47 (24.22) & 4.16 (12.35) & 1.43 (3.28) & 7.82 (25.33) & 5.14 (15.29) & 2.13 (7.41) & 1.16 (1.82) \\
CLIP & 20.35 (60.62) & 16.58 (41.25) & 5.21 (21.28) & 1.54 (5.56) & 14.28 (45.18) & 11.31 (30.12) & 3.14 (14.22) & 1.27 (3.18) & 9.17 (30.16) & 7.18 (18.11) & 1.56 (8.12) & 1.08 (1.52) \\
Fine-tuned CLIP & 20.06 (58.76) & 15.16 (39.84) & 6.08 (22.95) & 1.76 (5.82) & 13.82 (43.21) & 10.43 (28.17) & 4.07 (15.11) & 1.41 (3.59) & 8.87 (28.19) & 6.42 (16.21) & 2.18 (9.14) & 1.19 (1.97) \\
Fine-tuned SigLIP & 23.41 (63.26) & 18.28 (44.31) & 8.42 (25.18) & 1.52 (7.82) & 16.56 (48.11) & 12.87 (33.22) & 5.86 (18.81) & 1.93 (5.16) & 11.18 (33.65) & 8.41 (20.17) & 3.27 (11.18) & 1.46 (2.81) \\
Fine-tuned BioCLIP & \textbf{26.17} (\textbf{66.42}) & \textbf{21.18} (\textbf{48.21}) & \textbf{11.23} (\textbf{28.14}) & \textbf{2.87} (\textbf{9.11}) & \textbf{19.12} (\textbf{52.19}) & \textbf{15.18} (\textbf{37.51}) & \textbf{7.82} (\textbf{21.18}) & \textbf{1.81} (\textbf{6.87}) & \textbf{13.48} (\textbf{37.11}) & \textbf{10.56} (\textbf{24.12}) & \textbf{4.83} (\textbf{14.16}) & \textbf{1.87} (\textbf{3.92}) \\
\midrule
\multicolumn{13}{l}{\textbf{1\% of training set}} \\
\midrule
Fine-tuned ViT & 21.85 (58.41) & 17.31 (42.49) & 10.45 (27.42) & 3.32 (12.05) & 15.48 (42.17) & 11.82 (29.51) & 6.81 (18.13) & 1.87 (7.82) & 10.72 (28.18) & 7.46 (18.68) & 4.13 (11.19) & 1.81 (4.13) \\
CLIP & 20.18 (60.45) & 16.52 (41.81) & 6.98 (24.42) & 1.78 (8.71) & 14.07 (44.18) & 11.17 (29.35) & 4.19 (16.43) & 0.93 (5.11) & 9.16 (29.78) & 6.82 (18.16) & 2.84 (9.81) & 1.38 (2.87) \\
Fine-tuned CLIP & 22.21 (61.22) & 16.65 (43.02) & 8.92 (26.18) & 3.98 (11.65) & 16.18 (45.11) & 11.52 (30.18) & 5.82 (17.81) & 1.72 (7.58) & 10.82 (30.34) & 7.18 (19.12) & 3.19 (10.82) & 1.86 (3.81) \\
Fine-tuned SigLIP & 25.28 (64.18) & 20.22 (47.48) & 12.31 (30.49) & 5.08 (14.83) & 18.33 (48.68) & 14.07 (34.47) & 8.16 (21.84) & 2.82 (9.98) & 12.16 (33.67) & 9.19 (22.53) & 5.33 (13.07) & 1.48 (5.92) \\
Fine-tuned BioCLIP & \textbf{28.17} (\textbf{67.14}) & \textbf{23.94} (\textbf{51.26}) & \textbf{15.81} (\textbf{34.01}) & \textbf{7.40} (\textbf{17.63}) & \textbf{21.04} (\textbf{52.35}) & \textbf{16.32} (\textbf{38.44}) & \textbf{10.66} (\textbf{25.55}) & \textbf{4.56} (\textbf{12.81}) & \textbf{15.67} (\textbf{37.93}) & \textbf{11.64} (\textbf{26.30}) & \textbf{6.58} (\textbf{16.22}) & \textbf{2.50} (\textbf{7.02}) \\
\midrule
\multicolumn{13}{l}{\textbf{10\% of training set}} \\
\midrule
Fine-tuned ViT & 30.71 (69.45) & 24.82 (53.05) & 17.25 (38.28) & 10.85 (24.95) & 22.41 (52.18) & 17.08 (38.37) & 11.39 (26.33) & 6.41 (16.23) & 15.16 (36.24) & 11.05 (25.36) & 6.82 (16.30) & 3.18 (9.76) \\
CLIP & 22.01 (62.31) & 17.52 (44.38) & 8.68 (27.14) & 4.32 (13.98) & 15.18 (45.12) & 12.07 (31.18) & 5.14 (18.27) & 2.76 (8.54) & 10.93 (30.25) & 7.82 (20.40) & 2.86 (11.01) & 1.82 (4.31) \\
Fine-tuned CLIP & 27.88 (67.22) & 22.05 (50.25) & 14.72 (35.81) & 10.35 (23.75) & 19.34 (49.30) & 15.38 (36.18) & 9.12 (24.16) & 5.46 (15.80) & 12.86 (34.35) & 9.82 (23.87) & 5.05 (14.77) & 2.82 (8.44) \\
Fine-tuned SigLIP & 34.76 (72.36) & 28.16 (56.80) & 20.52 (42.00) & 13.99 (28.32) & 25.76 (55.43) & 20.67 (41.06) & 14.91 (30.47) & 8.01 (19.24) & 18.35 (40.14) & 14.16 (28.64) & 9.02 (19.82) & 4.67 (11.49) \\
Fine-tuned BioCLIP & \textbf{38.16} (\textbf{76.96}) & \textbf{32.31} (\textbf{60.24}) & \textbf{24.85} (\textbf{47.70}) & \textbf{16.13} (\textbf{32.86}) & \textbf{29.23} (\textbf{59.27}) & \textbf{24.02} (\textbf{45.37}) & \textbf{17.14} (\textbf{34.82}) & \textbf{11.31} (\textbf{23.38}) & \textbf{21.32} (\textbf{44.06}) & \textbf{17.62} (\textbf{32.82}) & \textbf{11.31} (\textbf{23.50}) & \textbf{6.19} (\textbf{14.57}) \\
\midrule
\multicolumn{13}{l}{\textbf{100\% of training set}} \\
\midrule
Fine-tuned ViT & 41.25 (75.46) & 33.15 (60.25) & 25.60 (45.84) & 17.16 (31.77) & 28.09 (58.24) & 22.51 (44.87) & 16.64 (31.23) & 9.32 (21.62) & 20.85 (42.88) & 15.16 (31.10) & 10.59 (21.22) & 5.14 (13.79) \\
CLIP & 24.99 (66.21) & 20.11 (48.13) & 11.88 (31.56) & 8.67 (17.16) & 18.25 (49.12) & 14.24 (34.81) & 7.43 (21.66) & 3.48 (11.58) & 12.40 (33.02) & 9.43 (23.15) & 4.07 (14.62) & 1.28 (6.13) \\
Fine-tuned CLIP & 42.31 (78.77) & 35.16 (64.83) & 28.25 (50.34) & 23.77 (36.02) & 31.03 (61.81) & 25.27 (48.05) & 18.30 (35.16) & 11.55 (24.79) & 23.35 (45.50) & 18.94 (35.54) & 12.25 (24.12) & 6.16 (15.27) \\
Fine-tuned SigLIP & 42.92 (78.06) & 36.86 (66.43) & 31.41 (57.05) & 28.42 (48.47) & 
39.19 (73.52) & 31.34 (60.94) & 25.73 (50.33) & 23.17 (41.85) & 
37.51 (71.59) & 23.30 (59.76) & 23.43 (46.63) & 18.29 (38.37) \\
Fine-tuned BioCLIP & \textbf{46.41} (\textbf{81.12}) & \textbf{40.30} (\textbf{69.00}) & \textbf{37.70} (\textbf{63.28}) & \textbf{30.26} (\textbf{52.07}) & 
\textbf{41.67} (\textbf{78.09}) & \textbf{38.69} (\textbf{68.81}) & \textbf{30.96} (\textbf{52.63}) & \textbf{27.15} (\textbf{46.01}) & 
\textbf{40.42} (\textbf{77.20}) & \textbf{31.64} (\textbf{64.12}) & \textbf{28.27} (\textbf{48.95}) & \textbf{20.60} (\textbf{41.76}) \\
\bottomrule

\end{tabular}
}
\end{table*}

\begin{table*}[ht]
\centering
\caption{Overall model evaluation. Results represent the mean accuracy (\%) $\pm$ standard deviation (\%) across all frequency categories over 5 random runs. The experiments using 100\% of training set do not include a standard deviation.}
\label{result_all}
\small
\setlength{\tabcolsep}{4pt}
\renewcommand{\arraystretch}{1.1}
\resizebox{\textwidth}{!}{
\begin{tabular}{c*{8}{c}}
\toprule
\multirow{2}{*}{\textbf{Model}} & \multicolumn{4}{c}{\textbf{Top-1}} & \multicolumn{4}{c}{\textbf{Top-5}} \\
\cmidrule(lr){2-5}\cmidrule(lr){6-9}
 & Order & Family & Genus & Species & Order & Family & Genus & Species \\
\midrule
\multicolumn{9}{l}{\textbf{0.1\% of training set}} \\
\midrule
Fine-tuned ViT & 16.37 $\pm$ 0.58 & 12.39 $\pm$ 0.30 & 5.84 $\pm$ 0.88 & 1.73 $\pm$ 0.22 & 49.15 $\pm$ 1.24 & 32.27 $\pm$ 0.22 & 18.49 $\pm$ 0.64 & 5.23 $\pm$ 0.55 \\
CLIP & 19.01 $\pm$ 0.28 & 15.43 $\pm$ 1.12 & 4.76 $\pm$ 1.28 & 1.48 $\pm$ 0.56 & 57.15 $\pm$ 0.70 & 38.71 $\pm$ 1.03 & 19.72 $\pm$ 1.46 & 5.05 $\pm$ 0.89 \\
Fine-tuned CLIP & 18.69 $\pm$ 0.44 & 14.12 $\pm$ 1.37 & 5.63 $\pm$ 0.19 & 1.69 $\pm$ 0.42 & 55.27 $\pm$ 0.66 & 37.20 $\pm$ 2.34 & 21.24 $\pm$ 0.87 & 5.34 $\pm$ 0.54 \\
Fine-tuned SigLIP & 21.91 $\pm$ 1.55 & 17.09 $\pm$ 0.18 & 7.84 $\pm$ 1.42 & 1.58 $\pm$ 0.67 & 59.86 $\pm$ 0.58 & 41.74 $\pm$ 1.46 & 23.70 $\pm$ 1.12 & 7.23 $\pm$ 0.69 \\
Fine-tuned BioCLIP & \textbf{24.62} $\pm$ 0.79 & \textbf{19.87} $\pm$ 0.94 & \textbf{10.47} $\pm$ 0.38 & \textbf{2.67} $\pm$ 0.91 & \textbf{63.18} $\pm$ 0.85 & \textbf{45.71} $\pm$ 1.06 & \textbf{26.57} $\pm$ 1.32 & \textbf{8.58} $\pm$ 0.44 \\
\midrule
\multicolumn{9}{l}{\textbf{1\% of training set}} \\
\midrule
Fine-tuned ViT & 20.46 $\pm$ 0.54 & 16.11 $\pm$ 0.46 & 9.66 $\pm$ 0.67 & 3.04 $\pm$ 0.27 & 54.82 $\pm$ 0.92 & 39.63 $\pm$ 0.74 & 25.40 $\pm$ 0.42 & 11.11 $\pm$ 0.65 \\
CLIP & 18.84 $\pm$ 0.33 & 15.34 $\pm$ 1.02 & 6.40 $\pm$ 0.63 & 1.63 $\pm$ 0.82 & 56.84 $\pm$ 0.64 & 39.04 $\pm$ 1.35 & 22.66 $\pm$ 1.82 & 7.94 $\pm$ 0.72 \\
Fine-tuned CLIP & 20.87 $\pm$ 0.66 & 15.52 $\pm$ 0.70 & 8.24 $\pm$ 0.85 & 3.55 $\pm$ 0.36 & 57.63 $\pm$ 0.62 & 40.18 $\pm$ 0.48 & 24.33 $\pm$ 0.34 & 10.74 $\pm$ 1.32 \\
Fine-tuned SigLIP & 23.74 $\pm$ 0.61 & 18.87 $\pm$ 0.78 & 11.42 $\pm$ 0.49 & 4.60 $\pm$ 0.85 & 60.70 $\pm$ 1.03 & 44.58 $\pm$ 2.31 & 28.53 $\pm$ 1.04 & 13.76 $\pm$ 0.79 \\
Fine-tuned BioCLIP & \textbf{26.62} $\pm$ 1.22 & \textbf{22.31} $\pm$ 1.32 & \textbf{14.68} $\pm$ 0.44 & \textbf{6.78} $\pm$ 0.96 & \textbf{63.81} $\pm$ 0.76 & \textbf{48.39} $\pm$ 0.68 & \textbf{32.07} $\pm$ 1.18 & \textbf{16.51} $\pm$ 0.58 \\
\midrule
\multicolumn{9}{l}{\textbf{10\% of training set}} \\
\midrule
Fine-tuned ViT & 28.87 $\pm$ 0.42 & 23.13 $\pm$ 1.57 & 15.97 $\pm$ 0.68 & 9.89 $\pm$ 0.32 & 65.59 $\pm$ 0.80 & 49.79 $\pm$ 0.56 & 35.64 $\pm$ 1.52 & 23.05 $\pm$ 0.85 \\
CLIP & 20.55 $\pm$ 0.44 & 16.33 $\pm$ 1.14 & 7.92 $\pm$ 0.76 & 3.99 $\pm$ 0.62 & 58.50 $\pm$ 0.32 & 41.48 $\pm$ 1.33 & 25.19 $\pm$ 0.38 & 12.79 $\pm$ 0.36 \\
Fine-tuned CLIP & 26.02 $\pm$ 0.55 & 20.58 $\pm$ 0.49 & 13.50 $\pm$ 0.84 & 9.32 $\pm$ 0.31 & 63.27 $\pm$ 2.33 & 47.13 $\pm$ 0.91 & 33.25 $\pm$ 1.55 & 21.97 $\pm$ 0.68 \\
Fine-tuned SigLIP & 32.78 $\pm$ 0.76 & 26.50 $\pm$ 1.63 & 19.24 $\pm$ 1.20 & 12.72 $\pm$ 0.85 & 68.59 $\pm$ 0.79 & 53.35 $\pm$ 1.43 & 39.42 $\pm$ 0.76 & 26.31 $\pm$ 1.31 \\
Fine-tuned BioCLIP & \textbf{36.18} $\pm$ 1.34 & \textbf{30.50} $\pm$ 1.06 & \textbf{23.17} $\pm$ 0.84 & \textbf{15.03} $\pm$ 0.61 & \textbf{73.05} $\pm$ 1.42 & \textbf{56.96} $\pm$ 0.76 & \textbf{44.84} $\pm$ 0.39 & \textbf{30.74} $\pm$ 0.57 \\
\midrule
\multicolumn{9}{l}{\textbf{100\% of training set}} \\
\midrule
Fine-tuned ViT & 38.47 & 30.85 & 23.67 & 15.51 & 71.63 & 56.83 & 42.69 & 29.55 \\
CLIP & 23.50 & 18.82 & 10.91 & 7.59 & 62.38 & 45.18 & 29.42 & 15.90 \\
Fine-tuned CLIP & 39.88 & 33.04 & 26.13 & 21.23 & 74.96 & 61.18 & 47.05 & 33.54 \\
Fine-tuned SigLIP & 42.14 & 35.53 & 30.24 & 27.25 & 77.12 & 65.33 & 55.63 & 47.08 \\
Fine-tuned BioCLIP & \textbf{45.45} & \textbf{39.76} & \textbf{36.31} & \textbf{29.45} & \textbf{80.50} & \textbf{68.81} & \textbf{61.10} & \textbf{50.76} \\
\bottomrule
\end{tabular}
}
\end{table*}

\paragraph{\textbf{Comparative analysis of model performance}}
Our comparative analysis reveals distinct performance dynamics among the models, heavily influenced by both the architecture and data scale. Across all data scales, the specialized vision models, particularly Fine-tuned BioCLIP and Fine-tuned SigLIP, consistently demonstrated superior performance compared to general-purpose architectures like Fine-tuned ViT and the zero-shot CLIP. Even in the extremely low-data scenario (0.1\% subset), Fine-tuned BioCLIP exhibited powerful few-shot generalization, achieving 26.17\% and 21.18\% Top-1 accuracy at the order and family levels for the \textit{frequent} class, outperforming all other models. As the training data scaled up to 100\%, finetuned-bioclip maintained its significant lead, achieving the highest accuracies across all taxonomic levels and frequency categories.

\paragraph{\textbf{Impact of data scale on model performance}}
Model performance on our task is highly dependent on the training set volume. As the training set size increases, all models show consistent improvement. For instance, the Top-1 species accuracy of Fine-tuned BioCLIP on the \textit{frequent} subset steadily climbs from 2.87\% (on the 0.1\% subset) to 7.40\% (1\%), 16.13\% (10\%) and 30.26\% (100\%). Crucially, the substantial performance jump from the 10\% subset to the 100\% subset indicates that model performance is far from saturated. This trend underscores the necessity of large-scale datasets, suggesting that further expanding the training data could continue to yield significant performance gains.

\paragraph{\textbf{Fine-grained nature of the classification task}}
The difficulty of the task is highlighted by the large gap between Top-1 and Top-5 accuracies, even when using the full training set. For example, the best-performing Fine-tuned BioCLIP model (at 100\% scale, \textit{frequent} class) achieved high Top-5 accuracies of 81.12\% (order), 69.00\% (family), 63.28\% (genus), and 52.07\% (species). In sharp contrast, its Top-1 accuracies were significantly lower at 46.41\%, 40.30\%, 37.70\%, and 30.26\% respectively. This discrepancy reveals the fine-grained nature of the challenge: models can effectively narrow down the possibilities, but they struggle to make the final, precise decision among visually similar taxonomic groups.

\paragraph{\textbf{Impact of class frequency on model performance}}
Our analysis reveals that model efficacy is heavily influenced by the sample distribution, with performance exhibiting a strict hierarchy across the \textit{frequent}, \textit{common}, and \textit{rare} classes. Due to the abundance of training samples, models generally demonstrate the strongest capability to learn representative features from the \textit{frequent} category, which acts as a relatively simpler task. Using the full training set, the Top-1 species accuracy of Fine-tuned BioCLIP strictly follows the \textit{frequent} > \textit{common} > \textit{rare} pattern, dropping from 30.26\% (\textit{frequent}) to 27.15\% (\textit{common}) and ultimately to 20.60\% (\textit{rare}). Similar degradation is consistently observed across other models, including Fine-tuned SigLIP (28.42\% > 23.17\% > 18.29\%) and Fine-tuned ViT (17.16\% > 9.32\% > 5.14\%). This trend highlights a fundamental challenge in fine-grained classification: while models can comfortably master well-represented classes, their generalization ability degrades significantly on \textit{common} and particularly \textit{rare} species, underscoring the bottleneck imposed by long-tailed class imbalance.

\section{Discussion}
Our expanded evaluation reveals a nuanced performance dynamic contingent on data scale. At lower data scales, the single-modality Fine-tuned ViT remains highly competitive, often performing comparably to or even outperforming Fine-tuned CLIP in several taxonomic categories. This suggests that with limited data, ViT can efficiently extract visual supervisory signals. However, a significant shift occurs when scaling to the full training set: the fine-tuned cross-modal models (Fine-tuned CLIP, Fine-tuned SigLIP, and particularly Fine-tuned BioCLIP) comprehensively surpass Fine-tuned ViT. This highlights the latent power of large-scale vision-language pretraining, which seemingly requires massive data to fully bridge the modal gap and adapt to fine-grained task. Specifically, Fine-tuned BioCLIP, which incorporates biological and taxonomic knowledge during pretraining, demonstrates the strongest capability across almost all scales, proving that domain-specific cross-modal alignment is highly effective for parsing the complex visual hierarchy of tree species. Although the four-level taxonomic labels used as supervision currently carry limited semantic information, the superiority of cross-modal architectures at scale is evident. In future work, incorporating large language models (LLMs) to generate more structured and semantically rich, human-readable descriptions of tree species, thereby replacing simple label concatenation with stronger semantic signals, may further unlock the potential of these models~\cite{hou2025urban, feng2025citybench}. However, even when trained on the full dataset with the best-performing BioCLIP, the Top-1 accuracy remains insufficient for practical deployment. This limitation arises not only from the models’ capacity but also from the inherent complexity of real-world urban tree classification: many species exhibit high inter-species visual similarity, while the same species can show substantial intra-species variation across seasons, cities, and climatic conditions; in addition, street view imagery is strongly affected by lighting, occlusion, camera viewpoint, and scene noise~\cite{huang2025no}, which introduces considerable uncertainty into visual appearance. In fact, even tree experts would find it difficult to make fully accurate species-level judgments from a single street view image in the absence of multi-view information or key details. Therefore, the limited Top-1 accuracy is a faithful reflection of the intrinsic difficulty of the task, and also indicates that current vision models still face clear performance bottlenecks when performing fine-grained classification in diverse urban environments.

The StreetTree dataset itself plays an important role in advancing urban ecological research and urban governance. StreetTree provides unprecedented global coverage, rich seasonal information, multi-temporal records, and fine-grained four-level taxonomic labels, offering a high-quality, easily accessible, and well-structured data foundation for building more powerful computer vision and deep learning models, as well as a standardized benchmark for large-scale tree-species classification. Compared with previous tree datasets that were often fragmented, limited in scale, or lacking temporal depth, StreetTree has higher information density and more comprehensive content, enabling researchers to conduct large-scale experiments more conveniently and systematically improve model performance in complex urban environments. Such data-driven approaches have the potential to substantially reduce the labor and time costs of urban tree inventory, health monitoring, and fine-grained management, and to promote a transition toward more automated, scalable, and intelligent urban forest management. The dataset will be made publicly available after the review process is completed.

Despite these advantages, the StreetTree dataset still has several limitations: the distribution of tree species across global cities is highly uneven, and data from some regions remain sparse. Future work may further enrich data sources, improve the accuracy of taxonomic and geographic annotations, and incorporate multimodal information to enhance the completeness and reliability of the dataset; explore more advanced visual models and multimodal large language models to improve fine-grained feature representation and cross-modal alignment; and develop methods to mitigate the impact of long-tailed distributions and seasonal fluctuations, while strengthening research on model interpretability. Ultimately, applying StreetTree in real-world urban forest management and biodiversity monitoring scenarios may help drive the development of urban ecological governance.

\section{Conclusion}

This study introduces \textbf{StreetTree}, the largest and most comprehensive global dataset of urban street trees to date, together with the rigorously designed evaluation set \textbf{StreetTree-18kEval}, providing a high-quality foundation for global research on fine-grained tree species classification. The more than 12 million samples in StreetTree span 133 countries, offering complete four-level taxonomic annotations, rich seasonal information, and long-term temporal records, thereby greatly enhancing the availability and representativeness of data for urban tree research. Through evaluations conducted on ViT, CLIP, SigLIP and BioCLIP models, we identify the key challenges of large-scale, fine-grained, and real-world tree classification, offering valuable insights for future model development and methodological advances. Furthermore, StreetTree is posed to drive progress in urban forest monitoring: with its high-resolution, multi-season, and spatially aligned global street view imagery, the dataset provides a robust basis for more accurate, scalable, and automated urban tree inventory and ecological analysis. We believe that StreetTree will serve as a crucial bridge between computer vision, deep learning, and urban ecology, supporting the future development of data-driven urban green management.

%
%
\bibliographystyle{splncs04}
\bibliography{main}

\title{Supplementary Material} 

\titlerunning{StreetTree}

\author{Jiapeng Li\inst{1} \and
Yingjing Huang\inst{2} \and
Fan Zhang\inst{3} \and
Yu Liu\inst{4}}

\authorrunning{Jiapeng L., Yingjing H., et al.}

\institute{Peking University,
\email{jpli25@stu.pku.edu.cn}\\
\and
University of Vienna, 
\email{yingjing.huang@univie.ac.at}\\
\and
Peking University,
\email{fanzhanggis@pku.edu.cn}\\
\and
Peking University,
\email{liuyu@urban.pku.edu.cn}\\
}

\maketitle

\section{StreetTree dataset statistics}
The StreetTree dataset is a large-scale and fine-grained global tree species classification resource. This chapter presents a systematic and in-depth description of its statistical properties and structural components.

\subsection{Basic statistics}
StreetTree offers georeferenced tree observations accompanied by a complete four-level taxonomic hierarchy and other structured attributes. The scale and coverage of the dataset are summarized as follows:
\begin{itemize}
    \item Total samples: 12,235,152
    \item Countries coverage: 133
    \item Taxonomic span: 71 orders, 241 families, 1,747 genera, and 8,363 species
    \item Data source distribution: U.S. 5 Million (44.58\%), GBIF (36.95\%), SGTrees (16.20\%), UrbanForest (1.52\%), Cambridge (0.73\%), and TreeML (0.01\%)
\end{itemize}

\subsection{Long tail distribution analysis}
\begin{figure*}[htbp]
  \centering
  \includegraphics[width=\textwidth]{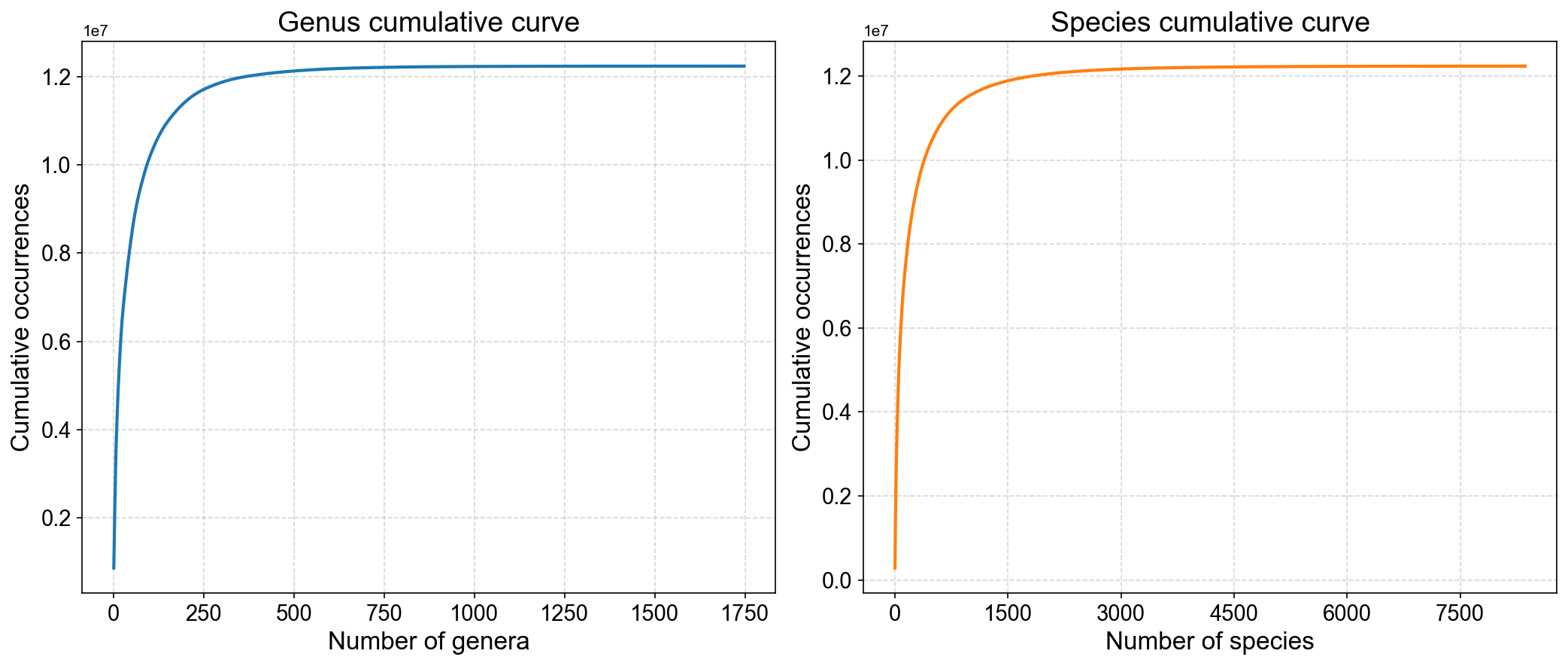}
  \caption{Long-tail distribution across genus and species level in StreetTree dataset.}
  \label{longtail}
\end{figure*}

The dataset exhibits a pronounced long-tailed distribution at both the genus and species levels, as illustrated in Figure~\ref{longtail}. At the genus level, the top 20\% of genera (349 in total) account for 97.85\% of all samples. In contrast, 460 genera contain no more than 10 samples each, forming a substantial long-tail of rare categories.

A similar pattern emerges at the species level: the top 20\% of species (1,672 in total) contribute 97.70\% of the dataset, whereas 3,333 species have 10 samples or fewer. This highlights an even more extreme long-tailed distribution at the species level.

This long-tailed distribution reflects high global ecological imbalance in the abundance of urban tree species, implying that models tend to learn features of frequent species more readily while exhibiting limited recognition performance on rare species. StreetTree-18kEval explicitly accounts for the impact of such uneven species distributions: by stratifying species according to their frequency category and applying balanced sampling, the evaluation set achieves both representativeness and uniformity across frequency categories.

\subsection{Detailed statistics of category attributes}
This chapter consolidates the statistical characteristics of the StreetTree dataset across several key attributes, encompassing its geographic coverage as well as the composition and diversity of its four-level taxonomic structure. Together, these analyses provide a comprehensive depiction of the dataset’s overall distributional landscape.

\paragraph{\textbf{Geographic coverage}}
The 20 most represented countries in the dataset are (ordered by descending sample count):
\begin{itemize}
    \item United States of America: 7,519,125
    \item Singapore: 1,993,414
    \item France: 521,830
    \item Australia: 395,828
    \item Italy: 222,089
    \item Spain: 207,863
    \item Chinese Taipei: 173,021
    \item Canada: 144,023
    \item New Zealand: 114,313
    \item Mexico: 95,923
    \item United States: 88,724
    \item United Kingdom of Great Britain and Northern Ireland: 88,666
    \item South Africa: 73,120
    \item Belgium: 59,099
    \item Portugal: 55,300
    \item Brazil: 52,429
    \item Colombia: 29,234
    \item India: 27,565
    \item Austria: 24,243
    \item Chinese Hong Kong: 24,026
\end{itemize}

\paragraph{\textbf{Distribution of order}} The 20 most represented orders in the dataset are (ordered by descending sample count):
\begin{itemize}
    \item Rosales: 2,131,752
    \item Sapindales: 1,403,450
    \item Fabales: 1,143,302
    \item Fagales: 1,092,558
    \item Myrtales: 1,091,736
    \item Lamiales: 825,348
    \item Pinales: 714,276
    \item Malpighiales: 613,183
    \item Malvales: 459,764
    \item Arecales: 459,568
    \item Proteales: 455,447
    \item Ericales: 274,870
    \item Saxifragales: 240,854
    \item Magnoliales: 221,303
    \item Laurales: 179,834
    \item Ginkgoales: 174,899
    \item Gentianales: 145,449
    \item Cornales: 134,571
    \item Apiales: 98,443
    \item Aquifoliales: 64,232
\end{itemize}

\paragraph{\textbf{Distribution of family}} The 20 most represented families in the dataset are (ordered by descending sample count):
\begin{itemize}
    \item Rosaceae: 1,310,149
    \item Fabaceae: 1,141,182
    \item Fagaceae: 902,601
    \item Sapindaceae: 743,156
    \item Myrtaceae: 726,603
    \item Oleaceae: 519,348
    \item Pinaceae: 486,376
    \item Arecaceae: 459,489
    \item Platanaceae: 424,147
    \item Anacardiaceae: 383,108
    \item Ulmaceae: 343,501
    \item Malvaceae: 328,931
    \item Lythraceae: 321,563
    \item Moraceae: 299,475
    \item Salicaceae: 265,510
    \item Meliaceae: 223,891
    \item Altingiaceae: 219,598
    \item Magnoliaceae: 208,388
    \item Lauraceae: 177,590
    \item Euphorbiaceae: 176,214
\end{itemize}

\paragraph{\textbf{Distribution of genus}} The 20 most represented genera in the dataset are (ordered by descending sample count):
\begin{itemize}
    \item Quercus: 869,243
    \item Acer: 504,763
    \item Prunus: 484,830
    \item Pyrus: 453,671
    \item Platanus: 424,147
    \item Fraxinus: 398,134
    \item Pinus: 377,435
    \item Lagerstroemia: 261,608
    \item Ulmus: 242,584
    \item Gleditsia: 234,843
    \item Liquidambar: 219,598
    \item Samanea: 212,010
    \item Ficus: 206,882
    \item Ginkgo: 174,798
    \item Tilia: 173,150
    \item Populus: 166,993
    \item Pistacia: 163,672
    \item Magnolia: 147,844
    \item Eucalyptus: 145,752
    \item Washingtonia: 137,736
\end{itemize}

\paragraph{\textbf{Distribution of species}} The 20 most represented species in the dataset are (ordered by descending sample count):
\begin{itemize}
    \item Pyrus calleryana: 395,414
    \item Platanus acerifolia: 283,858
    \item Samanea saman: 208,287
    \item Gleditsia triacanthos: 204,578
    \item Lagerstroemia indica: 187,583
    \item Liquidambar styraciflua: 137,077
    \item Acer rubrum: 137,007
    \item Peltophorum pterocarpum: 132,352
    \item Fraxinus pennsylvanica: 125,066
    \item Acer platanoides: 125,010
    \item Quercus palustris: 117,743
    \item Washingtonia robusta: 116,794
    \item Syagrus romanzoffianum: 113,552
    \item Prunus cerasifera: 113,319
    \item Tilia cordata: 106,212
    \item Zelkova serrata: 100,486
    \item Ginkgo biloba: 96,838
    \item Quercus virginiana: 94,330
    \item Magnolia grandiflora: 91,926
    \item Rhaphiolepis bibas: 83,902
\end{itemize}

\subsection{Real-world challenges and complexities in the dataset}
Although the StreetTree dataset has undergone extensive cleaning and quality check to achieve best performance, it still confronted the substantial complexities inherent to real-world street view images. In the raw Google Street View (GSV) images, trees are frequently affected by a range of uncontrolled visual factors, including occlusions caused by street infrastructures, excessively strong or weak lighting conditions, intra-species variability driven by seasonal changes, tree misalignment arising from imprecise camera angles and visually ambiguous boundaries among highly similar species. These issues collectively increase the difficulty of data filtering process and pose new challenges to model training.
Below, we provide representative examples of these issues in the raw data to illustrate the underlying challenges involved in the StreetTree dataset.

\begin{figure}[htbp]
  \centering
  \includegraphics[width=0.8\linewidth]{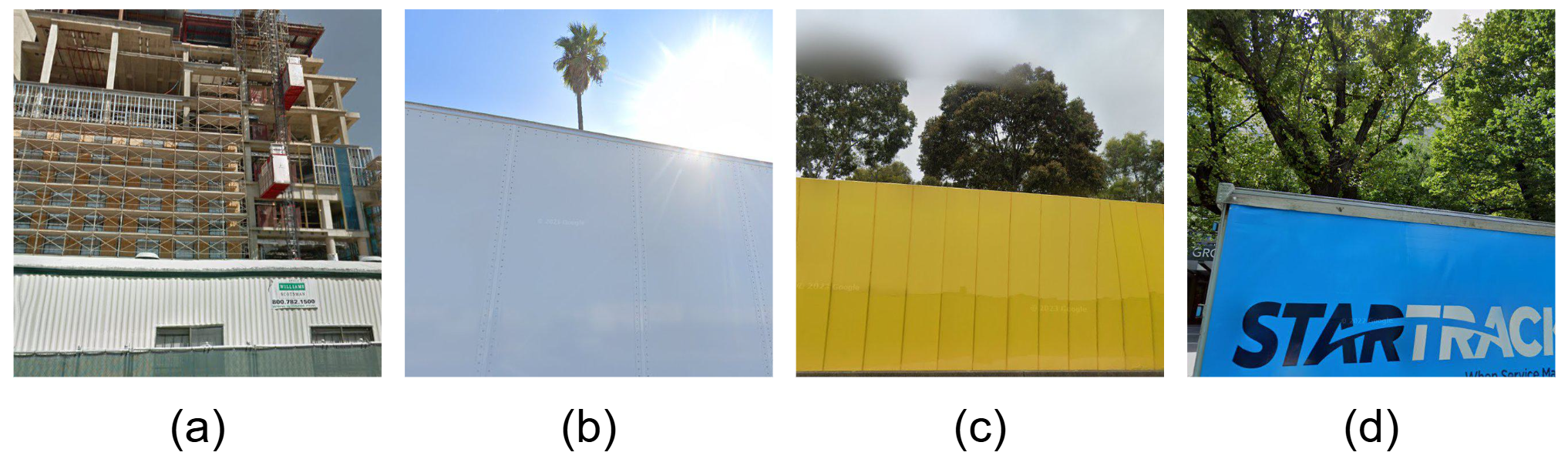}
  \caption{Example of occlusions caused by street infrastructures.}
  \label{occlutions}
\end{figure}

\paragraph{\textbf{Occlusions caused by street infrastructures}}
Tree crowns in street view images are frequently blocked by buildings, traffic signs or other urban structures, resulting in incomplete or visually distorted observations. Such occlusions make it difficult for models to capture the full morphological characteristics of the tree. A portion of such data has already been filtered out during the data cleaning process, and Figure~\ref{occlutions} presents several examples.

\begin{figure}[htbp]
  \centering
  \includegraphics[width=0.8\linewidth]{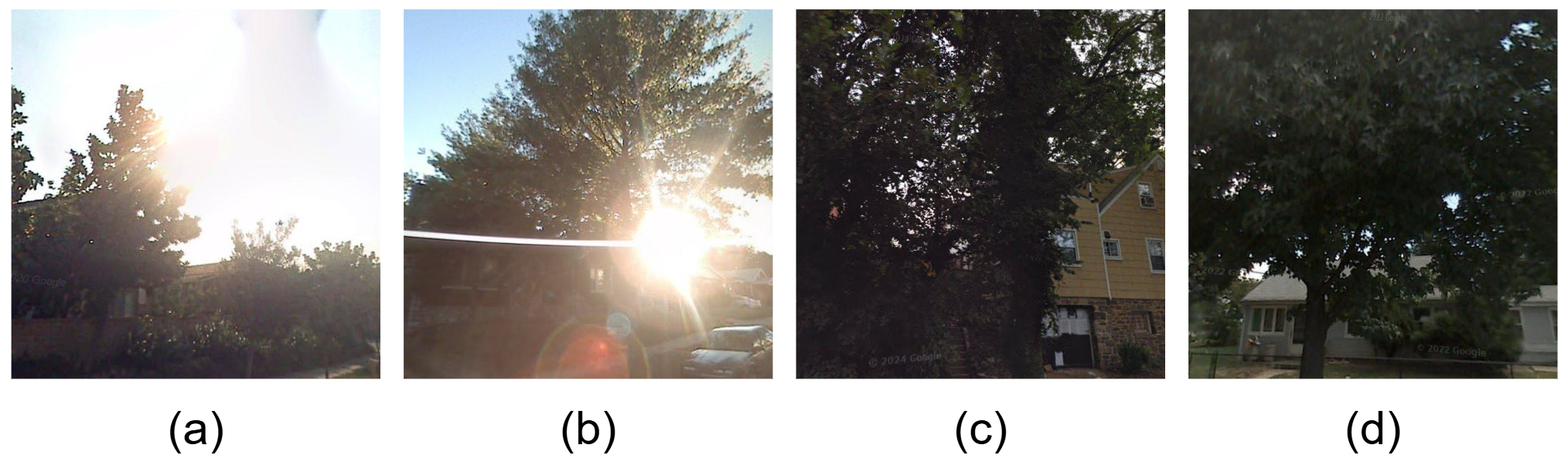}
  \caption{Example of excessively strong or weak lighting conditions. \textbf{(a)} and \textbf{(b)} illustrate cases of excessively strong lighting, while \textbf{(c)} and \textbf{(d)} show cases of insufficient illumination.}
  \label{lighting}
\end{figure}

\paragraph{\textbf{Excessively strong or weak lighting conditions}}
Extreme lighting, either overly bright or insufficient, can obscure fine-grained visual cues such as leaf texture or branch structure. These illumination variations introduce substantial noise into feature extraction and species recognition. Figure~\ref{lighting} presents several examples.

\begin{figure}[htbp]
  \centering
  \includegraphics[width=0.8\linewidth]{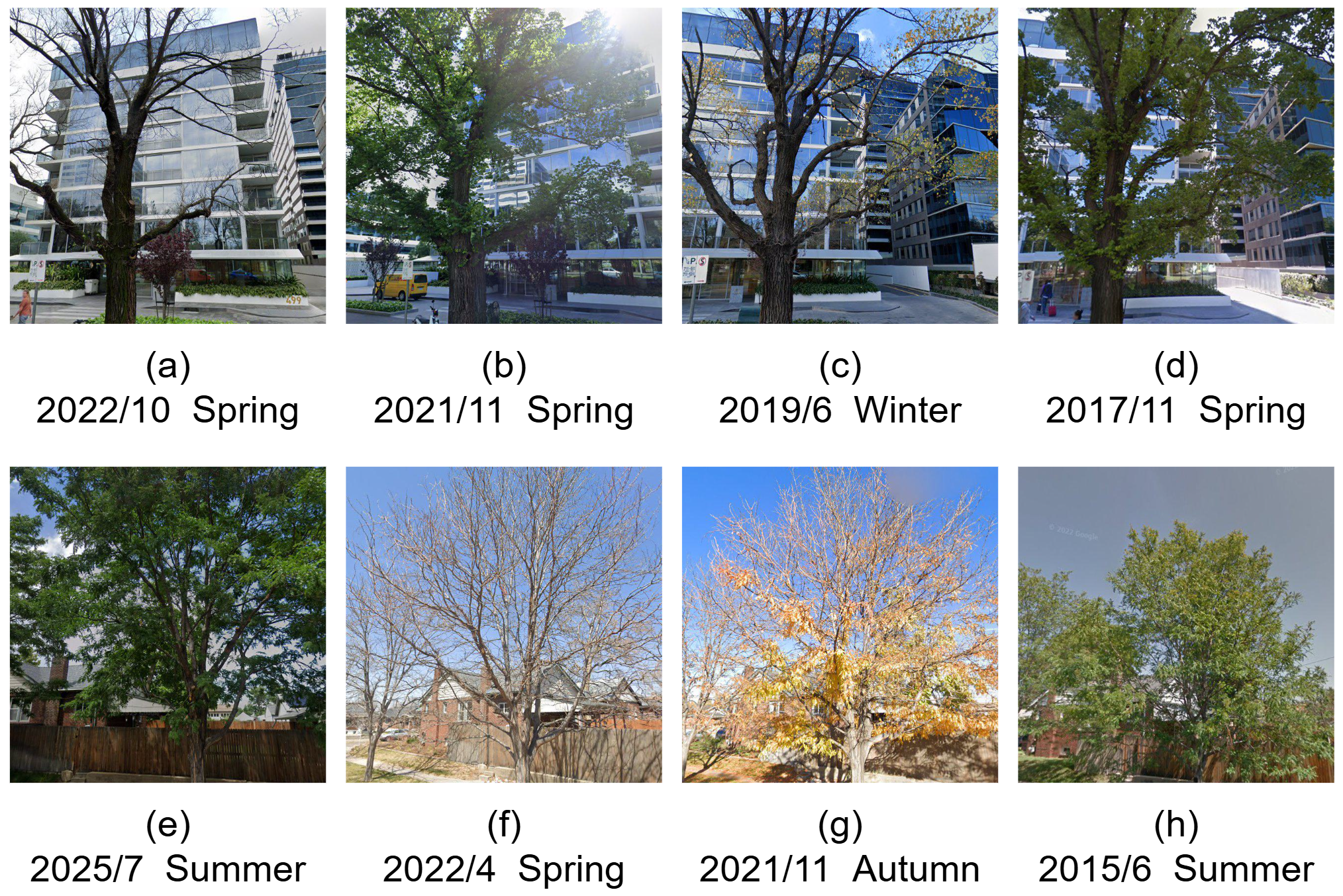}
  \caption{Example of intra-species variability driven by seasonal changes. \textbf{(a)-(d)} and \textbf{(e)-(h)} illustrate the seasonal progression observed in two individual trees, illustrating their seasonal variability. \textbf{(a)} corresponds to early spring, where the tree has recently emerged from dormancy and the branches remain bare. \textbf{(b)} and \textbf{(d)} show the tree in mid to late spring, characterized by dense and deep green leaves. \textbf{(c)} appears to depict early winter or a late autumn state where most leaves have fallen, leaving only a few yellow leaves clinging to the branches. For another tree, a similar pattern is posed. \textbf{(e)} and \textbf{(h)} were captured in summer, showing full, vibrant green leaves. \textbf{(f)} represents winter, during which the branches are completely bare. \textbf{(g)} reflects an autumn condition, with abundant yellow leaves signaling seasonal transition.}
  \label{seasonal}
\end{figure}

\paragraph{\textbf{Intra-Species variability driven by seasonal changes}}
Seasonal transitions cause dramatic changes, like Figure~\ref{seasonal}. Difference in crown and breach morphology, leaf color, and overall tree appearance can leading to large intra-species variability, which complicates the learning of stable discriminative features. 

\begin{figure}[htbp]
  \centering
  \includegraphics[width=0.8\linewidth]{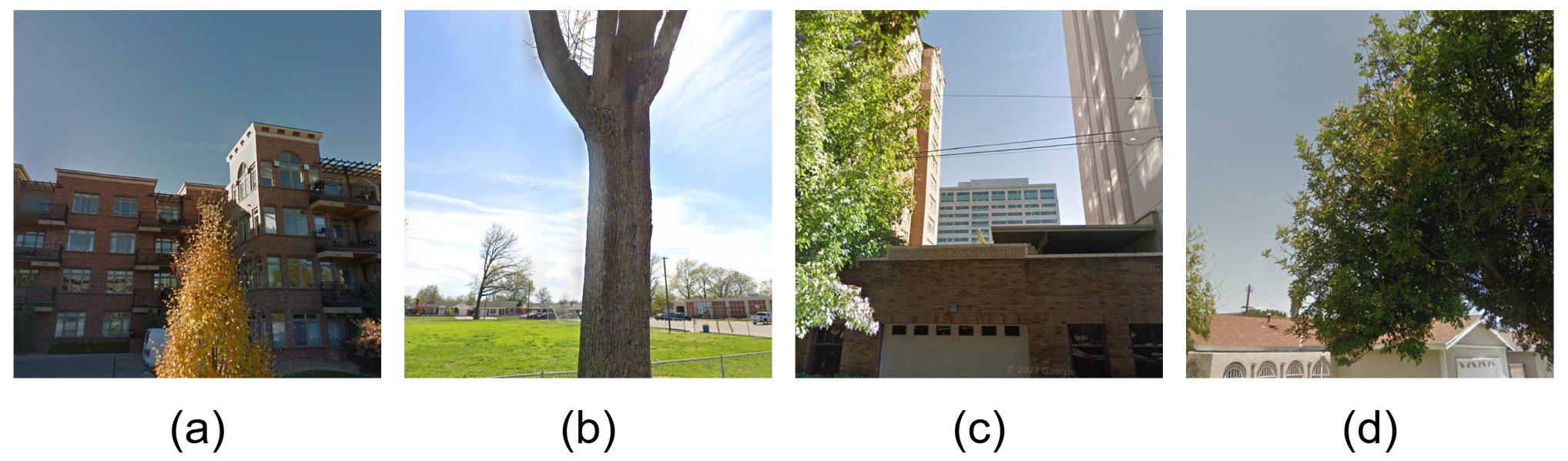}
  \caption{Example of tree misalignment arising from imprecise camera angles. \textbf{(a)-(d)} illustrate how inaccurate camera angle cause the target tree to appear too low, too high, too left, or too right in the captured image respectively.}
  \label{camera}
\end{figure}

\paragraph{\textbf{Tree misalignment arising from imprecise camera angles}}
Inaccuracies in GSV camera orientation or GPS alignment may capture the wrong tree or only part of intended target. These misalignments reduce labeling precision and introduce unintended visual context, see Figure~\ref{camera}. 

\begin{figure}[htbp]
  \centering
  \includegraphics[width=0.8\linewidth]{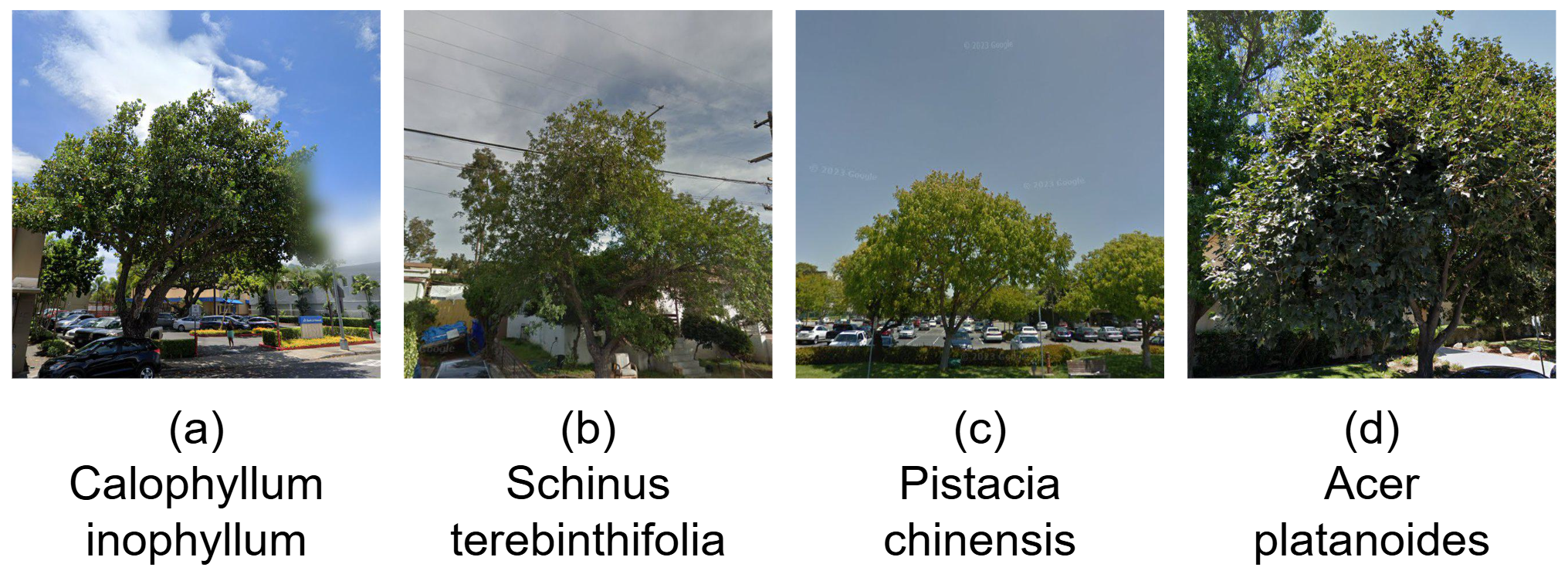}
  \caption{Example of visually ambiguous boundaries among highly similar species. The species name of each tree is shown below corresponding image.}
  \label{similar}
\end{figure}

\paragraph{\textbf{Visually ambiguous boundaries among highly similar species}}
Many tree species share extremely similar visual traits, leading to ambiguous boundaries even in high-quality images. Such fine-grained similarities pose inherent challenges for reliable species-level classification. Figure~\ref{similar} presents several species of visually similar yet taxonomically different.

\section{Further details of evaluation set}
To enable fair and representative benchmarking across different taxonomic levels and species, we constructed an independent evaluation set, StreetTree-18kEval. We first partitioned all species in the full StreetTree dataset into three frequency categories: frequent, common, and rare, based on their observed sample counts. From each frequency group, we then randomly selected 30 species. Finally, we randomly sampled 500, 100, and 20 images from the frequent, common, and rare species respectively, thereby forming an evaluation set with a controlled gradient of species frequency. This design enables rigorous assessment of a model’s generalization ability across species with markedly different frequency categories.
Table~\ref{evaluationset} below summarizes the taxonomic composition of the evaluation set.

\begin{table*}[ht]
\centering
\caption{Taxonomic composition of StreetTree-18kEval dataset.}
\label{evaluationset}
\small
\setlength{\tabcolsep}{4pt}
\renewcommand{\arraystretch}{1.1}
\resizebox{\textwidth}{!}{
\begin{tabular}{c*{12}{c}}
\toprule
\textbf{Order} & \textbf{Family} & \textbf{Genus} & \textbf{Species} & \textbf{Sample count} \\
\midrule
\multicolumn{5}{l}{\textbf{Frequent species}} \\
\midrule
Pinales & Pinaceae & Picea & Picea pungens & 500 \\
Rosales & Rosaceae & Malus & Malus sylvestris & 500 \\
Laurales & Lauraceae & Machilus & Machilus japonica & 500 \\
Fagales & Fagaceae & Quercus & Quercus ilicifolia & 500 \\
Myrtales & Myrtaceae & Melaleuca & Melaleuca linariifolia & 500 \\
& & ... & & \\
\textbf{Total} & & (A total of 30 frequent species were selected) & & 15,000 \\
\midrule
\multicolumn{5}{l}{\textbf{Common species}} \\
\midrule
Malpighiales & Salicaceae & Salix & Salix bonplandiana & 100 \\
Fabales & Fabaceae & Dalbergia & Dalbergia cochinchinensis & 100 \\
Malvales & Malvaceae & Durio & Durio zibethinus & 100 \\
Magnoliales & Myristicaceae & Knema & Knema globularia & 100 \\
Lamiales & Oleaceae & Notelaea & Notelaea microcarpa & 100 \\
& & ... & & \\
\textbf{Total} & & (A total of 30 common species were selected) & & 3,000 \\
\midrule
\multicolumn{5}{l}{\textbf{Rare species}} \\
\midrule
Aquifoliales & Aquifoliaceae & Ilex & Ilex micrococca & 20 \\
Ericales & Ericaceae & Kalmia & Kalmia latifolia & 20 \\
Celastrales & Celastraceae & Euonymus & Euonymus europaeus & 20 \\
Laurales & Lauraceae & Nectandra & Nectandra oppositifolia & 20 \\
Dipsacales & Viburnaceae & Viburnum & Viburnum opulus & 20 \\
& & ... & & \\
\textbf{Total} & & (A total of 30 rare species were selected) & & 600 \\
\bottomrule
\end{tabular}
}
\end{table*}

\section{Algorithm for azimuth angle calculation}
After obtaining the geographic coordinates of each tree, we retrieve the corresponding street view images through the GSV API. Since GSV camera usually captures panoramic imagery centered at the capture point, it is essential to specify an angle that directs the camera toward the target tree. To achieve this, we first search for available GSV capture locations around each tree and compute the azimuth angle from the capture point to the tree, which is then used as the heading parameter $\alpha$ in the GSV API.

Specifically, let the capture point be located at $(\varphi_1, \lambda_1)$ and the tree at $(\varphi_2, \lambda_2)$, where $\varphi$ denotes latitude and $\lambda$ denotes longitude. Let $\Delta\lambda = \lambda_2 - \lambda_1$. The azimuth angle $\theta$ from the capture point to the tree is computed as:
\[
\theta = \operatorname{atan2}\!\left(
\begin{aligned}
&\sin(\Delta\lambda)\cos\varphi_2,\\[-0.3em]
&\cos\varphi_1\sin\varphi_2
 - \sin\varphi_1\cos\varphi_2\cos(\Delta\lambda)
\end{aligned}
\right)
\]

Since $\theta$ lies in the range $(-\pi, \pi]$, we convert it to degrees and map it to $[0, 360)$ to obtain the final heading $\alpha$:
\[
\alpha =
\left(
\theta \times \frac{180}{\pi} + 360
\right) \bmod 360
\]

By determining an appropriate GSV capture point and its corresponding azimuth angle for each tree, we are able to precisely control the camera orientation and ensure that the tree appears within the main field of image. This procedure enables large-scale acquisition of high-quality street view images and provides the foundation for subsequent data-cleaning steps.

\section{Data availability}
The StreetTree dataset constructed in this study is derived entirely from publicly accessible sources, including TreeML, U.S. 5 Million, UrbanForest, Singapore Trees, GBIF API, and other open data sources. All data comply with open licenses (CC0 1.0, CC BY 4.0, CC BY-NC 4.0) and contain no personally identifiable or sensitive information, ensuring that the dataset adheres to ethical and legal standards throughout its use.
The dataset will be made publicly available upon completion of the review process. The StreetTree-12M training set and the StreetTree-18kEval evaluation set will be hosted on the HuggingFace platform to facilitate efficient access for the community. We are committed to the continued maintenance and expansion of the StreetTree dataset, including addressing user feedback, improving data quality, and integrating additional high-value data sources to further support research on urban trees.

Following standard practices in the computer vision community for large-scale web-sourced datasets, the StreetTree dataset is released as a comprehensive metadata index rather than a collection of raw image files. Specifically, the released dataset includes \textbf{the unique panorama identifier (Panorama ID, PanoID), geographical coordinates (latitude and longitude), the corresponding observation time (year and month), the local season, the country of origin, complete multi-level taxonomic information (Order, Family, Genus, and Species), and the species frequency category (\textit{frequent}, \textit{common}, or \textit{rare})}. To access the visual data, researchers can fetch the images directly via the official Google Street View API. This distribution format ensures full reproducibility and serves as a lightweight, tree-centric index to facilitate future urban ecological research.

\end{document}